\RequirePackage{fix-cm}
\documentclass[smallextended]{svjour3}       
\smartqed  
\usepackage{graphicx}
\usepackage{amsmath}
\usepackage{dsfont}
\usepackage{booktabs}
\usepackage{multirow}
\usepackage{mathrsfs}
\usepackage{amssymb}
\usepackage{color}
\usepackage{subfigure}
\usepackage[ruled]{algorithm2e}
\usepackage[colorlinks,linkcolor=blue]{hyperref}

\begin{document}

\title{Binarized Neural Architecture Search for Efficient Object Recognition
}

\titlerunning{BNAS}        

\author{Hanlin Chen \and Li'an Zhuo \and Baochang Zhang* \and Xiawu Zheng \and Jianzhuang Liu  \and Rongrong Ji  \and David Doermann \and Guodong Guo
}

\authorrunning{H. Chen, L. Zhuo, B. Zhang} 

\institute{Hanlin Chen, Li'an Zhuo, Baochang Zhang* (corresponding author) \at
              Beihang University, Beijing \\
              \email{\{hlchen, lianzhuo, bczhang\}@buaa.edu.cn}
           \and
           Xiawu Zheng, Rongrong Ji \at
              Xiamen University, Fujian \\
              \email{\{zhengxiawu, rrji\}@buaa.edu.cn}
            \and
           Jianzhuang Liu, David Doermann, Guodong Guo are respectively with \at
              Shenzhen Institutes of Advanced Technology, Shenzhen, China; University at Buffalo, New York; Institute of Deep Learning, Institute of Deep Learning, Baidu Research;  National Engineering Laboratory for Deep Learning Technology and Application.  \\
              \email{jz.liu@siat.ac.cn; doermann@buffalo.edu; guoguodong01@baidu.com}
}

\date{Received: date / Accepted: date}

\maketitle

\begin{abstract}
		Traditional neural architecture search (NAS) has a significant impact in computer vision by automatically designing   network architectures for various tasks.
		In this paper, binarized neural architecture search (BNAS), with a search space of binarized convolutions, is introduced to produce extremely compressed models to reduce huge computational cost on embedded devices for edge computing. The BNAS calculation is more challenging than NAS due to the learning inefficiency caused by optimization requirements and the huge architecture space, and the performance loss when handling the wild data in various computing applications. To address these issues, we introduce operation space reduction and channel sampling into BNAS to significantly reduce the cost of searching. This is accomplished through a performance-based strategy that is robust to wild data, which is further used to  abandon less potential operations. Furthermore, we introduce the Upper Confidence Bound (UCB) to solve  1-bit BNAS. Two optimization methods for binarized neural networks are used to validate the effectiveness of our BNAS. Extensive experiments demonstrate that the proposed BNAS achieves a comparable performance  to NAS on both CIFAR and ImageNet databases. An accuracy of $96.53\%$ vs. $97.22\%$ is achieved on the CIFAR-10 dataset, but with a significantly compressed model, and a $40\%$ faster search than the state-of-the-art PC-DARTS. On the wild face recognition task, our binarized models achieve a performance similar to their corresponding full-precision models.
\end{abstract}

\section{Introduction}
\label{intro}

    Efficient computing has become one of the hottest topics both in academy and industry. It will be vital for the $5$G networks by providing hardware-friendly and efficient solutions for practical and wild applications \cite{mao2017mobile}. Edge computing is about computing resources that are closer to the end user. This makes applications faster and users friendly \cite{chen2019deep}. It enables mobile or embedded devices to provide real-time intelligent analysis of big data, which can  reduce the pressure on the cloud computing center and improve the availability \cite{han2019convergence}. However, edge computing is still challenged by its limited computational ability, memory and storage and severe performance loss, making the models for edge computing inefficient for feature calculation and inference \cite{li2019edge}.  

    One possible solution for efficient edge computing can be achieved based on compressed deep models, which mainly fall into three lines: network pruning, knowledge distillation and model quantization. Network pruning \cite{han2015learning} aims to remove network connections with less significance, and knowledge distillation \cite{hinton2015distilling} introduces a teacher-student model, which uses the soft targets generated by the teacher model to guide the student model with much smaller model size, to achieve knowledge transfer. Differently, model quantization \cite{paper10} calculates neural networks with low-bit weights and activations to compress a model in a more efficient way, which is also orthogonal to the other two. The binarized model is widely considered as  one of the most efficient ways to perform computing on embedded devices with an extremely less computational cost. Binarized filters have been used in traditional convolutional neural networks (CNNs) to compress deep models \cite{rastegari2016xnor,paper10,paper14,paper15}, showing up to 58-time speedup and 32-time memory saving. In \cite{paper15}, the XNOR network is presented where both the weights and inputs attached to the convolution are approximated with binary values.  This results in an efficient implementation of convolutional operations by reconstructing the unbinarized filters with a single scaling factor. \cite{Zhuang_2018_CVPR} introduces $2\!\sim\!4$-bit quantization based on a two-stage approach to quantize the weights and activations, which significantly improves the efficiency and performance of quantized models. Furthermore, WAGE \cite{ICLR2018wu} is proposed to discretize both the training and inference processes, and it quantizes not only weights and activations, but also gradients and errors.  In \cite{gu2018projection}, a projection convolutional neural network (PCNN) is proposed to realize binarized neural networks (BNNs) based on a simple back propagation algorithm. In our previous work \cite{zhao2019bonn}, we  propose a novel approach, called Bayesian optimized 1-bit CNNs (denoted as BONNs), taking the advantage of Bayesian learning to significantly improve the performance of extreme 1-bit CNNs. There are also other practices in \cite{tang2017train,alizadeh2018empirical,ding2019regularizing} with improvements over previous works. Binarized models show the  advantages on computational cost reduction and memory saving, but they unfortunately suffer from  performance loss when handling wild data in practical applications. The main reasons are twofold. On the one hand, there is still a gap between low-bit weights/activations and full-precision weights/activations on  feature representation, which should be investigated from new perspectives. On the other hand, traditional binarized networks are based on the neural architecture manually designed for full-precision networks, which means that binarized architecture design remains largely unexplored. 
    
    \begin{figure*}[htbp!]
		\centering
		\includegraphics[scale=.22]{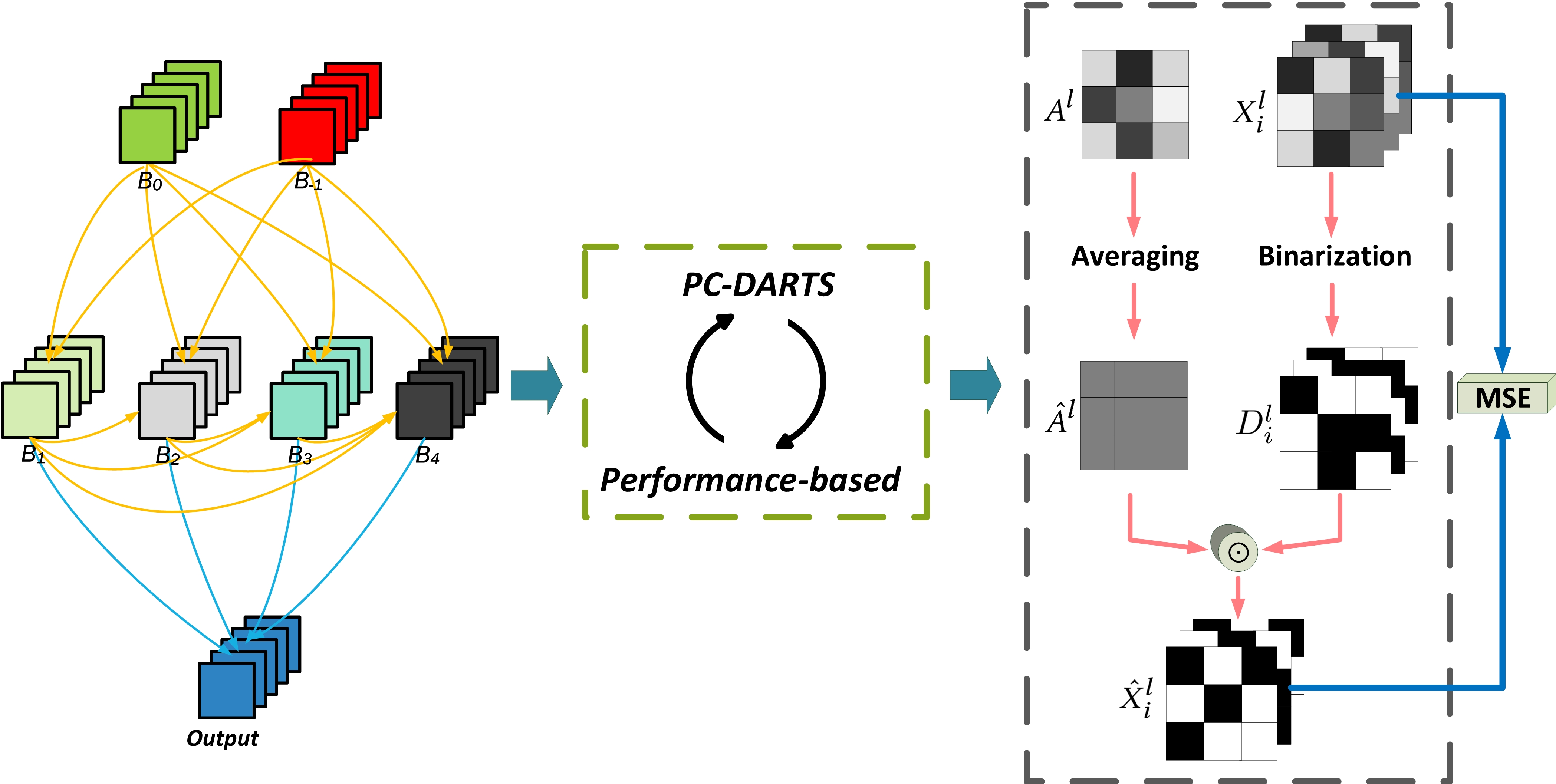}
		\caption{The overall framework of the proposed binarized neural architecture search (BNAS).  In BNAS,  the search cell is a fully connected directed acyclic graph with four nodes, which is calculated based on PC-DARTS and a performance-based method. We also reformulate the optimization of binarization of CNNs in the same framework.}
		\label{fig:bnas}
	\end{figure*}

	Traditional neural architecture search (NAS) has attracted great attention with a remarkable performance in various deep learning tasks. Impressive results have been shown for reinforcement learning (RL) based methods \cite{Zoph2018CVPR,zoph2016neural}, for example, which train and evaluate more than $20,000$ neural networks across $500$ GPUs over $4$ days. Recent methods like differentiable architecture search (DARTS) reduce the search time by formulating the task in a differentiable manner \cite{liu2018darts}. DARTS relaxes the search space to be continuous, so that the architecture can be optimized with respect to its validation set performance by gradient descent, which provides a fast solution for effective network architecture search. To reduce the redundancy in the network space,  partially-connected DARTS (PC-DARTS) was recently  introduced to perform a more efficient search without compromising the performance of DARTS \cite{xu2019pcdarts}.
	
	Although DARTS or its variants has a smaller model size than  traditional light models, the searched network still suffers from an inefficient  inference process due to the complicated  architectures generated by multiple stacked  full-precision convolution operations.  Consequently, the searched network for embedded device is still computationally expensive and inefficient. At the same time, the existing gradient-based approaches select operations without a meaningful guidance. Not only is the search process inefficient, but also the selected operation might exhibit significant vulnerability to model attacks based on gradient information \cite{goodfellow2014explaining,madry2017towards}, also for the wild data. Clearly, these problems require further exploration to overcome these challenges.
	
	To address these above challenges, we  transfer the NAS to a binarized neural architecture search (BNAS), by exploring the advantages of binarized neural networks (BNNs) on memory saving and computational cost reduction. In our BNAS framework as shown in Fig. \ref{fig:bnas}, we use PC-DARTS as a warm-up step, which is followed by the performance-based method to improve the robustness of the resulting BNNs for the wild data. In addition, based on the observation that the early optimal operation is not necessarily the optimal one in the end, and the worst operation in the early stage usually has a worse performance at the end \cite{zheng2019multinomial}.   We exploit the advantages of both PC-DARTS and performance  evaluation to prune the operation space. This means that the operations we finally reserve are certainly a near an optimal solution. On the other hand, with the operation pruning process, the search space becomes smaller and smaller, leading to an efficient search process.  We show that the BNNs obtained by BNAS can outperform conventional BNN models by a large margin. It is a significant contribution in the field of BNNs, considering that the performance of conventional BNNs are not yet comparable with their corresponding full-precision models in terms of accuracy.  To further validate the performance of our method, we also implement 1-bit BNAS in the same framework. Differently from BNNs (only kernels are binarized),  1-bit CNNs suffer from  poor performance evaluation problem for binarized operations with binarzied activations in the beginning due to the insufficient training. {We assume BNAS as a multi-armed bandit problem and introduce an exploration term based on the upper confidence bound (UCB) \cite{auer2002finite} to improve the search performance. 
The exploration term is used to handle the exploration-exploitation dilemma in the multi-armed bandit problem.} We lead a new performance measure based on UCB by considering both the performance evaluation and number of trial for operation pruning in the same framework, which means that the operation is ultimately abandoned only when it is sufficiently evaluated. 
	
	\begin{figure*}[htbp!]
		\centering
		\includegraphics[scale=.34]{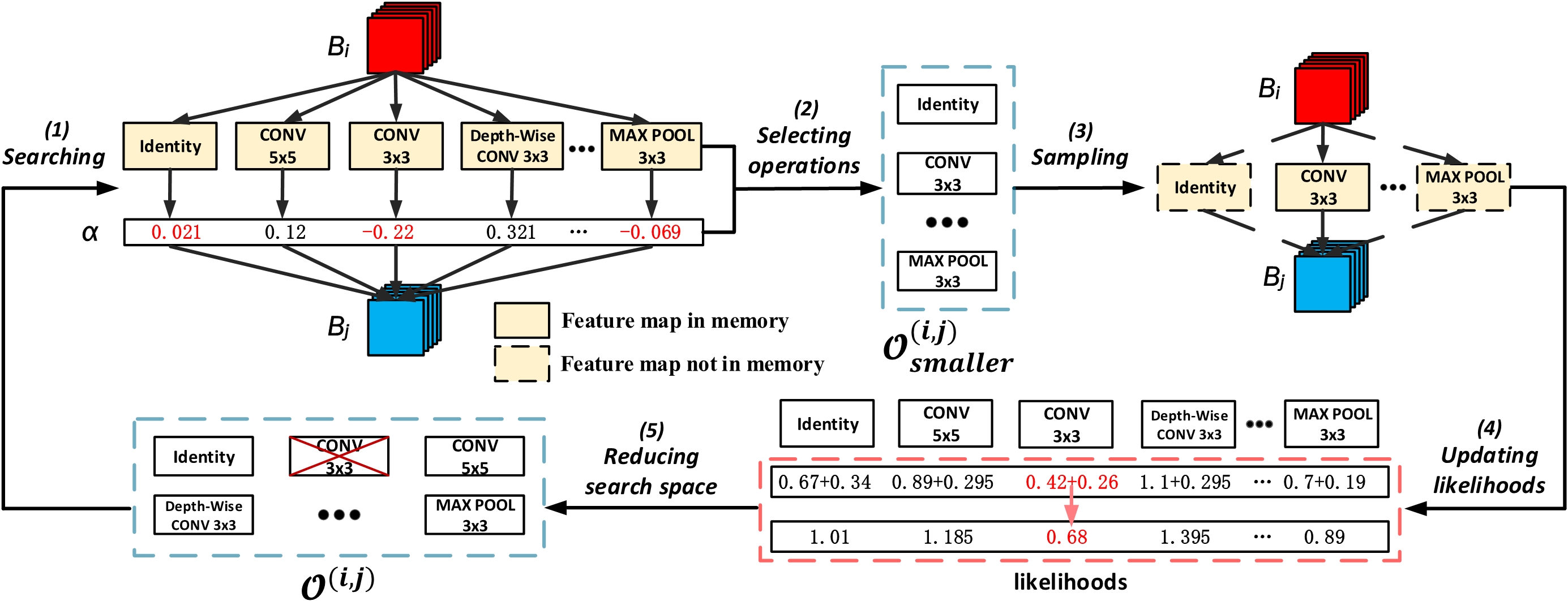}
		\caption{The main steps of our BNAS: (1) Search an architecture based on $\mathcal{O}^{(i,j)}$ using PC-DARTS. (2) Select half the operations with less potential from $\mathcal{O}^{(i,j)}$ for each edge, resulting in $\mathcal{O}^{(i,j)}_{smaller}$. (3) Select an architecture by sampling (without replacement) one operation from $\mathcal{O}^{(i,j)}_{smaller}$ for every edge, and then train the selected architecture. (4) Update the operation selection likelihood $s(o^{(i,j)}_k)$ based on the accuracy obtained from the selected architecture on the validation data. (5) Abandon the operation with the minimal selection likelihood from the search space $\{\mathcal{O}^{(i,j)}\}$ for every edge.}
		\label{fig:performance-based}
	\end{figure*}
	
	The search process of our BNAS consists of two steps. One is the operation potential ordering based on partially-connected DARTS (PC-DARTS) \cite{xu2019pcdarts} which also  serves as a baseline for our BNAS. It is further improved with a second operation reduction step guided by a performance-based strategy. 
	In the operation reduction step, we prune one operation at each iteration from one-half of the operations with less potential as calculated by PC-DARTS. As such, the optimization of the two steps becomes faster and faster because the search space is reduced due to the operation pruning. We can take advantage of the differential framework of DARTS where the search and performance evaluation are in the same setting. We also enrich the search strategy of DARTS. Not only is the gradient used to determine which operation is better, but the proposed performance evaluation is included for further reduction of the search space. The contributions of our paper include:
	
	\begin{itemize}
		\item
		BNAS is developed based on a new search algorithm which solves the BNNs and 1-bit CNNs optimization and architecture search in a unified framework. The  1-bit CNNs are obtained by incorporating the bandit strategy into BNAS, which can better evaluate the operation based on UCB.
		\item
		The search space is greatly reduced through a performance-based strategy used to abandon  operations with less potential, which improves the search efficiency by $40\%$.
		
		\item
		Extensive experiments demonstrate that the proposed algorithm achieves much better performance than other light models on  wild face recognition, CIFAR-10 and ImageNet. 
	\end{itemize}
	
	This submission is an extension of our conference paper \cite{aaai2020}  by including: 1)  extending our binarized models to 1-bit models, which are more challenging than BNNs;  In addition, the  1-bit CNNs are achieved based on the bandit strategy, which can better evaluate the operation based on UCB. 2) adding more details about optimization of binarized models; 3) adding more experiments to sufficiently validate the performance of our methods, such as new experiments on wild face recognition, and results  of 1-bit BNAS  on all the datasets.
	
	\section{Related Work}
	In this section, we introduce the most related works on  network quantization and NAS (DARTS). For the network quantization, both state-of-the-art BNNs and 1-bit CNNs are briefly introduced. We also described the PC-DARTS method, which are combined with binarized models, leading to a much better performance on object recognition tasks.  
    \subsection{Neural Networks Quantization} 

    To the best of our knowledge,  \cite{paper10} is the first attempt to binarize both the weights and activations of convolution layers in CNNs.
    It works well in maintaining the classification accuracy on small datasets like CIFAR-10 and CIFAR-100 \cite{krizhevsky2014cifar}, which is however less effective when being applied on large datasets like ImageNet \cite{rastegari2016xnor,deng2009imagenet}. \
    Instead of binarizing the kernel weights into {$\pm1$}, the work in \cite{rastegari2016xnor} adds a layer-wise scalar $\alpha_l$ to reconstruct the binarized kernels and proves that the mean absolute value (MAV) of each layer is the optimal value for $\alpha_l$. 
    Inspired by using a scalar to reconstruct binarized kernels, HQRQ \cite{li2017performance} adopts a high-order binarization scheme to achieve more accurate approximation while preserving the advantage of binary operation.
    In order to alleviate the degradation in prediction accuracy, ABC-Net \cite{lin2017towards} adopts multiple binary weights and activations to approximate full-precision weights. 
    \cite{leng2018extremely} decoupled the continuous parameters from the discrete constraints of network using ADMM, which therefore achieves extremely low bit rates. 
    Recently, Bi-real Net \cite{liu2018bi} explores a new variant of residual structure to preserve the real activations before the sign function, with a tight approximation to the derivative of the non-differentiable sign function. 
    \cite{mcdonnell2018training} applied a warm-restart learning-rate schedule to quantize network weights into 1-bit, which achieves about 98\%$\sim$99\% of peak performance on CIFAR.

    Quantizing kernel weights and activations to binary values is an extreme case of neural network quantization, which is prone to unacceptable accuracy degradation. 
    Accordingly, sufficient attention has been paid to quantize DCNNs with more than 1 bit. 
    Specifically, ternary weights are introduced to reduce the quantization error in TWN \cite{TWN}. 
    DoReFa-Net \cite{zhou2016dorefa} exploits convolution kernels with low bit-width parameters and gradients to accelerate both the training and inference. 
    TTQ \cite{zhu2016trained} uses two full-precision scaling coefficients to quantize the weights to ternary values. 
    \cite{Zhuang_2018_CVPR} presented a $2\!\sim\!4$-bit quantization scheme using a two-stage approach to alternately quantize the weights and activations, which provides an optimal tradeoff among memory, efficiency and performance. 
    Furthermore, WAGE \cite{ICLR2018wu} is proposed to discretize both the training and inference processes, where not only weights and activations but also gradients and errors are quantized.   	
    Other practices are shown in \cite{tang2017train,alizadeh2018empirical,ding2019regularizing} with improvements over previous works.

    Despite the excellent efficiency, existing 1-bit CNNs suffer from its limited  representation capability, leading to an inevitable performance loss on the object recognition tasks.      
    Our previous works  \cite{gu2018projection,zhao2019bonn} have significantly improved the performance of
    state-of-the-art 1-bit CNNs. However, the performance are still baffled by their manually designed architectures, and this paper exploits the BNAS method to further enhance the capability of BNNs, aiming to significantly reduce the gap to their full-precision counterparts.

	\subsection{Neural Architecture Search}
	Thanks to the rapid development of deep learning, significant gains in performance have been realized in a wide range of computer vision tasks, most of which are manually designed network architectures \cite{krizhevsky2012imagenet,simonyan2014very,he2016deep,huang2017densely}. Recently, the new approach called neural architecture search (NAS) has been attracting increased attention. The goal is to find automatic ways of designing neural architectures to replace conventional hand-crafted ones. Existing NAS approaches need to explore a very large search space and can be roughly divided into three type of approaches: evolution-based, reinforcement-learning-based and one-shot-based.

	In order to implement the architecture search within a short period of time, researchers try to reduce the cost of evaluating each searched candidate. Early efforts include sharing weights between searched and newly generated networks \cite{cai2018efficient}.  Later, this method was generalized into a more elegant framework named one-shot architecture search \cite{brock2017smash,cai2018proxylessnas,liu2018darts,pham2018efficient,xie2018snas,zheng2019multinomial,zheng2019dynamic}. In these approaches, an over-parameterized network or super network covering all candidate operations is trained only once, and the final architecture is obtained by sampling from this super network. For example, \cite{brock2017smash} trained the over-parameterized network using a HyperNet \cite{2016Hypernetworks}, and \cite{pham2018efficient} proposed to share parameters among child models to avoid retraining each candidate from scratch. DARTS \cite{liu2018darts} introduces a differentiable framework and thus combines the search and evaluation stages into one. Despite its simplicity, researchers have found some of its drawbacks and proposed a few improved approaches over DARTS \cite{xie2018snas,Chen_2019_ICCV}.
	PDARTS \cite{Chen_2019_ICCV} presents an efficient algorithm which allows the depth of searched architectures to grow gradually during the training procedure, with a significantly reduced search time.
	ProxylessNAS \cite{cai2018proxylessnas} adopted
    the differentiable framework and proposed to search architectures
    on the target task instead of adopting the conventional
    proxy-based framework.

	Unlike previous methods, the calculation of BNAS is more challenging due to the learning inefficiency and huge architecture search space,  we implement BNAS based on combination of PC-DARTS and new performance measures. We prune one operation at each iteration from one-half of the operations with smaller weights calculated by PC-DARTS, and thus the search becomes faster and faster in the optimization. As such, BNAS shows stronger robustness to wild data than DARTS with gradient-based search strategy.

    \subsection{Bandit problem}
    {In probability theory, the multi-armed bandit problem is a problem in which a decision must be made among competing choices in a way that maximizes their expected gain.  Each choice's properties are only partially known at any given time, and may become better understood as time passes or observed after the choices. The selection and following observations  provide information useful in future choices. The aim is to minimize the distance from the optimal solution with the shortest time. A lot of breakthroughs have been made for the bandit problem for constructing the optimal selection policies with fastest rate of convergence \cite{lai1985bandit}.}
    
    {Bandit optimization is commonly used to exemplify the exploration-exploit-
    ation trade-off dilemma to avoid an explosive traversal space and speed up optimal convergence. The upper confidence bound applied to trees (UCT) was propoesd as a bandit based Monte Carlo planning \cite{lev2006mc}. It is also exploited to improve classical reinforcement learning methods such as Q-learning \cite{even2006qlearning} and  state–action–reward–state–action (SARSA) \cite{mic2011sarsa}.
    AlphaGo \cite{alphago} modifies the original UCB multi-armed bandit policy by approximately predicting good arms at the start of a sequence of multi-armed bandit trials, which is called PUCB (predictor of upper confidence bounded) to balance the result of simulation and its uncertainty.}
    
    {Objective functions for the multi-armed bandit problem tend to take one of two flavors: 1) best arm identification (or pure exploration) in which one is interested in identifying the arm with the highest average payoff, and 2) exploration-versus-exploitation in which one tries to maximize the cumulative payoff over time \cite{se2012bandit}. Many optimization problems are studied in non-stochastic setting as the pull of each arm without the i.i.d. assumption \cite{neu2015nonsto,li2018hyperband,Jam2015half}. Relatedly, hyperband \cite{li2018hyperband} solves the pure-exploration bandit problem in the fixed budget setting  without making parametric assumptions and achieves the state-of-the-art for the hyperparameter optimization. It extends the Successive Halving Algorithm \cite{Jam2015half} which evaluates and throws out the worst half until one remains. We share the similar idea of resources allocation with hyperband and formulate our BNAS as an exploration-versus-exploitation problem where the sampling and abandoning are based on UCB.}
	 
	\section{Binarized Neural Architecture Search}
	
	In this section, we first describe the search space in a general form, where the computation procedure for an architecture (or a cell in it) is represented as a directed acyclic graph. {We then describe binarized optimaization for BNAS and review the baseline PC-DARTS \cite{xu2019pcdarts}, which is used as warm-up for our method. Then an operation sampling and a performance-based search strategy are proposed to effectively reduce the search space. Our BNAS framework is shown in Fig. \ref{fig:performance-based} and additional details of it are described in the rest of this section. Finally, we reformulate the optimization of BNNs in a unified framework.} 
	
	\subsection{Search Space}
\begin{figure}[htbp]
		\centering
		\subfigure[Cell]{ 
			\includegraphics[scale=.5]{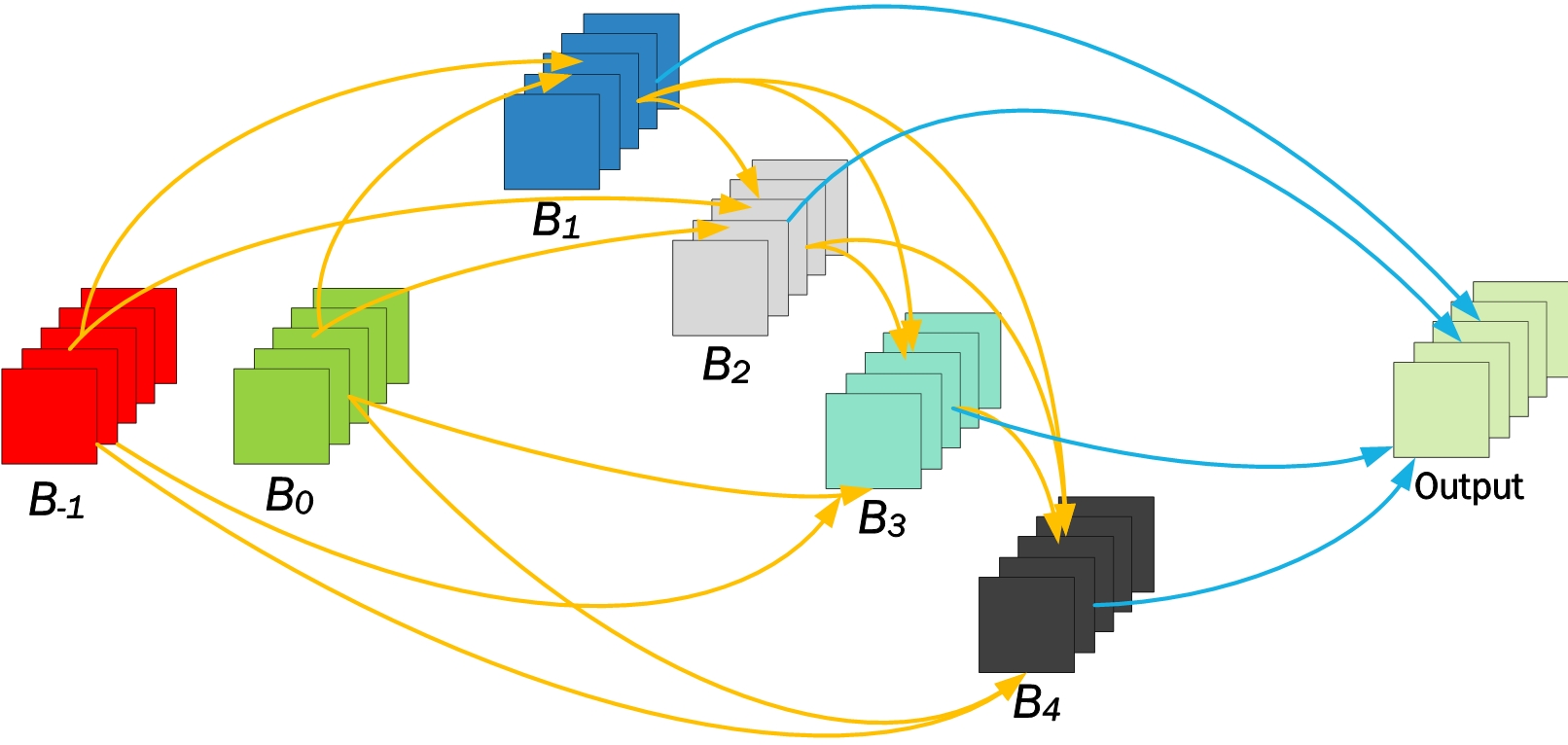}
			\label{fig:cell}
		}
		\quad
		
		\subfigure[Operation Set]{ 
			\includegraphics[scale=.5]{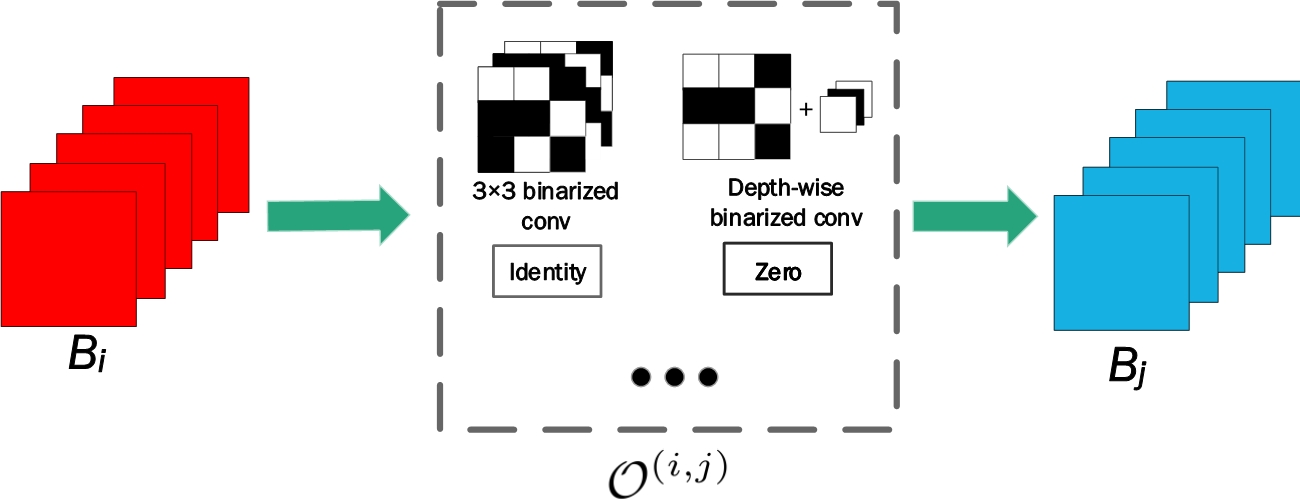}
			\label{fig:operationset}
		}
		\caption{ (a) A cell contains 7 nodes, two input nodes $B_{-1}$ and $B_0$, four intermediate nodes $B_1$, $B_2$, $B_3$, $B_4$ that apply sampled operations on the input nodes and upper nodes, and an output node that concatenates the outputs of the four intermediate nodes. (b) The set of operations $\mathcal{O}^{(i,j)}$ between $B_i$ and $B_j$, including binarized convolutions.}
	\end{figure}

	Following  \cite{zoph2016neural,Zoph2018CVPR,liu2018darts,real2019regularized}, we search for a computation cell as the building block of the final architecture. A network consists of a pre-defined number of cells \cite{zoph2016neural}, which can be either normal cells or reduction cells. Each cell takes the outputs of the two previous cells as input. A cell is a fully-connected directed acyclic graph (DAG) of $M$ nodes, \emph{i.e.}, $\{B_1, B_2, ..., B_M\}$, as illustrated in Fig. \ref{fig:cell}. Each node $B_i$ takes its dependent nodes as input, and generates an output through a sum operation $B_j = \sum_{i<j} o^{(i,j)}(B_i).$ Here each node is a specific tensor (\emph{e.g.,} a feature map in convolutional neural networks) and each directed edge $(i,j)$ between $B_i$ and $B_j$ denotes an operation $o^{(i,j)} (.)$, which is sampled from $\mathcal{O}^{(i,j)}=\{o^{(i,j)}_1, ..., o^{(i,j)}_K\}$. Note that the constraint $i<j$ ensures there are no cycles in a cell. Each cell takes the outputs of two dependent cells as input, and we define the two input nodes of a cell as $B_{-1}$ and $B_{0}$ for simplicity. Following \cite{liu2018darts}, the set of the operations $\mathcal{O}$ consists of $K = 8$ operations.  They include $3\times3$ max pooling, no connection (zero), $3\times3$ average pooling, skip connection (identity), $3\times3$ dilated convolution with rate $2$, $5\times5$ dilated convolution with rate $2$, $3\times3$ depth-wise separable convolution, and $5\times5$ depth-wise separable convolution, as illustrated in Fig. \ref{fig:operationset}. The search space of a cell is constructed by the operations of all the edges, denoted as $\{\mathcal{O}^{(i,j)}\}$.
	
	Unlike conventional convolutions, our BNAS is achieved by transforming all the convolutions in $\mathcal{O}$ to binarized convolutions. We denote the  full-precision and binarized kernels as $X$ and $\hat X$ respectively. A convolution operation in $\mathcal{O}$ is represented as  $B_j = B_i \otimes \hat X $ as shown in Fig. \ref{fig:operationset}, where $\otimes$ denotes convolution. To build BNAS, one key step is how to binarize the kernels from  $X$ to $\hat X$, which can be implemented based on state-of-the-art BNNs, such as XNOR or PCNN. As we know, the optimization of BNNs is more challenging than that of conventional CNNs \cite{gu2018projection,rastegari2016xnor}, which adds an additional burden to NAS. To solve it, we introduce channel sampling and  operation space reduction into differentiable NAS to significantly reduce the cost of GPU hours, leading to an efficient BNAS.

	
	\subsection{Binarized Optimization for BNAS}
	\label{opt}
	\begin{table*}[htbp]
		\caption{A brief description of the main notations used in section \ref{opt}.}
		\centering
		\begin{tabular}{l l l}
			\toprule
			$X$: full-precision kernel & $\hat{X}$: binarized kernel & $A$: amplitude matrix\\
			$F$: feature map & $D$: $X'$s direction & $\hat A$: generated from A\\
			\hline
			$i$: kernel index & $g$: input feature map index & $h$: output feature map index\\
			$S$: number of examples & $l$: layer index & $M$: number of facial landmarks\\
			\bottomrule
		\end{tabular}
		\label{notation}
	\end{table*}
	{The inference process of a BNN model is based on the binarized kernels, which means that the kernels must be binarized in the forward step (corresponding to the inference) during training. Contrary to the forward process, during back propagation, the resulting kernels are not necessary to be binarized and can be full-precision.}
	
	In order to achieve binarized weights, {we first divide each convolutional kernel into two parts (amplitude and direction)}, and formulate the current binarized methods in a unified framework.  In addition to Tab. \ref{notation}, we elaborate $D$, $A$ and $\hat A$: $D^l_i$ are the directions of the full-precision kernels $X^l_i$ of the $l^{th}$ convolutional layer, $l\in\lbrace1,\cdots,N\rbrace$; $A^l$ shared by all $D^l_i$ represents the amplitude of the $l^{th}$ convolutional layer; $\hat A^l$ and $A^l$ are of the same size and all the elements of $\hat A^l$ are equal to the average of the elements of $A^l$. In the forward pass, $\hat A^l$ is used instead of the full-precision $A^l$. In this case, $\hat A^l$ can be considered as a scalar.  The full-precision $A^l$ is only used for back propagation during training. Noted that our formulation can represent both XNOR based on scalar, and also simplified PCNN \cite{gu2018projection}  
whose scalar is learnable as a projection matrix.
	{We represent $\hat X$ by the amplitude and direction as}
		\begin{equation}
			\hat{X} = \hat A \odot D,
			\end{equation}
	where $\odot$ denotes the element-wise multiplication between matrices. We then define an amplitude loss function to reconstruct the full-precision kernels as
		
		\begin{equation}
			 	L_{\hat A} = \frac{\theta}{2}\sum_{i, l}\|X^l_i-\hat{X}^l_i\|^2 = \frac{\theta}{2}\sum_{i, l}\|X^l_i- \hat A^l \odot D^l_i\|^2,
		 	\label{loss_A}
		\end{equation}
	where $D_i^l=sign(X_i^l)$ represents the binarized kernel.  $X_i^l$ is the full-precision model which is updated during the back propagation process in PCNNs, while $\hat A^l$ is calculated based on a closed-form solution in XNOR. The element-wise multiplication combines the binarized kernels and the amplitude matrices to approximate the full-precision kernels. The final loss function is defined by considering
			\begin{equation}
		 	L_S = \frac{1}{2S}\sum_s\|\hat{Y}_s - Y_s\|^2_2,
			\end{equation}
	where $\hat{Y}_s$ is the label of the $s^{th}$ example; $Y_s$ is the corresponding classification results. Finally, the overall loss function $L$ is applied to supervise the training of BNAS in the  back propagation as
		\begin{equation}
		  	L = L_S + L_{\hat A}.
		  	\label{loss_L}
		\end{equation}

	{The binarized optimization is used to optimize the neural architecture search, leading to our binarized neural architecture search (BNAS).
	To this end,  we use partially-connected DARTS (PC-DARTS) to achieve operation potential ordering,
	which  serves as a warm-up step for our BNAS. Denote by $L_{train}$ and $L_{val}$ the training loss and the validation loss, respectively. Both losses are determined by not only the architecture $\alpha$ but also the binarized weights $\hat{X}$ in the network. The goal for the warm-up step is to find $\hat{X}^*$ and $\alpha^*$ that minimize the validation loss $L_{val}(\hat{X}^*, \alpha^*)$, where the weights $\hat{X}^*$ associated
    with the architecture are obtained by minimizing the training loss $\hat{X}^* = \mathop {\arg \min }\limits_{\hat{X}}\ L_{train}(\hat{X}, \alpha^*)$.
    }
    
    {This implies a bilevel optimization problem with $\alpha$ as the upper-level variable and $\hat{X}$ as the lower-level variable:
	\begin{equation}
    	\begin{aligned}
    	\mathop {\arg \min }\limits_{\alpha}\  
    	&L_{val}(\hat{X}^*, \alpha)\\
    	\textit{s.t.}\ &\hat{X}^* = \mathop {\arg \min }\limits_{\hat{X}}\ L_{train}(\hat{X}, \alpha).
    	\end{aligned}
	\end{equation}    
    }
    

	To better understand our method, we also review the core idea of PC-DARTS, which  can take advantage of partial channel connections to improve memory efficiency. Taking the connection from $B_i$ to $B_j$ for example, this involves defining a channel sampling mask $S^{(i,j)}$, which assigns $1$ to selected channels and $0$ to masked ones. The selected channels are sent to a mixed computation of $|\mathcal{O}^{(i,j)}|$ operations, while the masked ones bypass these operations. They are directly copied to the output, which is formulated as
	
	\begin{equation}
	\label{eq:pc}
	\begin{aligned}
	&f^{(i,j)}(B_i,S^{(i,j)}) \\
	&= \sum_{o^{i,j}_k\in \mathcal{O}^{(i,j)}} \frac{exp\{\alpha_{o^{(i,j)}_k}\}}{\sum_{o^{(i,j)}_{k^{'}} \in \mathcal{O}^{(i,j)}} exp\{\alpha_{o^{(i,j)}_{k^{'}}}\} } \cdot o^{(i,j)}_k(S^{(i,j)} * B_i)\\
	& + (1 - S^{(i,j)}) * B_i,
	\end{aligned}
	\end{equation}
	where $S^{(i,j)} * B_i$ and $(1 - S^{(i,j)}) * B_i$ denote the selected and masked channels, respectively, and $\alpha_{o^{(i,j)}_k}$ is the parameter of operation $o^{(i,j)}_k$ between $B_i$ and $B_j$.
	
	PC-DARTS sets the proportion of selected channels to $1/C$ by regarding $C$ as a hyper-parameter. In this case, the computation cost can also be reduced by $C$ times. However, the size of the whole search space is $2 \times K^{|\mathcal{E_M}|}$, where $\mathcal{E_M}$ is the set of possible edges with $M$ intermediate nodes in the fully-connected DAG, and the "$2$" comes from the two types of cells. In our case with $M=4$, together with the two input nodes, the total number of cell structures in the search space is $2 \times 8^{2+3+4+5} = 2 \times 8^{14}$.  This is an extremely large space to search for a binarized neural architectures which need more time than a full-precision NAS. Therefore, efficient optimization strategies for BNAS are required.
	

	\subsection{Performance-based Strategy for BNAS} \label{sec:PBS}
	Reinforcement learning is inefficient in the architecture search due to the delayed rewards in network training, i.e., the evaluation of a structure is usually done after the network training converges. On the other hand, we can perform the evaluation of a cell when training the network. Inspired by \cite{ying2019bench}, we use a performance-based strategy to boost the search efficiency by a large margin. \cite{ying2019bench} did a series of experiments showing that in the early stage of training, the validation accuracy ranking of different network architectures is not a reliable indicator of the final architecture quality. However, we observe that the experiment results actually suggest a nice property that if an architecture performs badly in the beginning of training, there is little hope that it can be part of the final optimal model. As the training progresses, this observation shows less uncertainty. Based on this observation, we derive a simple yet effective operation abandoning process. During training, along with the increasing epochs, we progressively abandon the worst performing operation in each edge.
	
	{To this end,
	we reduce the search space $\{\mathcal{O}^{(i,j)}\}$ after the warm-up step achieved by PC-DARTS to increase search efficiency. According to $\{\alpha_{o^{(i,j)}_k}\}$, we can select half of the operations with less potential from $\mathcal{O}^{(i,j)}$ for each edge, resulting in $\mathcal{O}^{(i,j)}_{smaller}$.}
	After that, we randomly sample one operation from the $K/2$ operations in $\mathcal{O}^{(i,j)}_{smaller}$ for every edge, then obtain the validation accuracy by training the sampled network for one epoch, and finally assign this accuracy to all the sampled operations. These three steps are performed $K/2$ times by sampling without replacement, leading to each operation having exactly one accuracy for every edge.
	
	We repeat it $T$ times. Thus each operation for every edge has $T$ accuracies $\{y_{k,1}^{(i,j)}, y_{k,2}^{(i,j)}, ..., y_{k,T}^{(i,j)}\}$. Then we define the selection likelihood of the $k$th operation in $\mathcal{O}^{(i,j)}_{smaller}$  for each edge as
	
	\begin{equation}\label{eq:performance_prob_smaller}
	s_{smaller}(o^{(i,j)}_k) = \frac{exp\{\bar{y}_k^{(i,j)}\}}{\sum_m exp\{\bar{y}_m^{(i,j)}\}},
	\end{equation}
	where $\bar{y}_k^{(i,j)} = \frac{1}{T}  \sum_t y_{k,t}^{(i,j)}$. And the selection likelihoods of the other operations not in $\mathcal{O}^{(i,j)}_{smaller}$ are defined as
	
	\begin{equation}\label{eq:performance_prob_larger}
	\begin{aligned}
	&s_{larger}(o^{(i,j)}_k) 
	=& \frac{1}{2} (\mathop{\max}\limits_{o^{(i,j)}_k}{\{s_{smaller}(o^{(i,j)}_k)\}} + \frac{1}{\lceil K/2 \rceil} \sum_{o^{(i,j)}_k}{s_{smaller}(o^{(i,j)}_k)}),
	\end{aligned}
	\end{equation}
	where $\lceil K/2 \rceil$ denotes the smallest integer $\geq K/2$. The reason to use it is because $K$ can be an odd integer during iteration  in the proposed Algorithm \ref{alg:MDL}. Eq. \ref{eq:performance_prob_larger} is an estimation for the rest operations using a  value balanced between the maximum and average of $s_{smaller}(o^{(i,j)}_k)$. Then, $s(o^{(i,j)}_k)$ is updated by 
	
	\begin{equation}\label{eq:performance_prob}
	\begin{aligned}
	s(o^{(i,j)}_k) \leftarrow & \frac{1}{2} s(o^{(i,j)}_k) + q_k^{(i,j)}s_{smaller}(o^{(i,j)}_k) + 
	& (1 - q_k^{(i,j)})s_{larger}(o^{(i,j)}_k),
	\end{aligned}
	\end{equation}
	where $q_k^{(i,j)}$ is a mask, which is $1$ for the operations in $\mathcal{O}_{smaller}^{(i,j)}$ and $0$ for the others. 
	
	When searching for BNAS, we do not use PC-DARTS as warm-up for the consideration of efficiency because quantizing feature maps is slower. Hence,  $\mathcal{O}^{(i,j)}_{smaller}$ is $\mathcal{O}^{(i,j)}$. Also, we introduce an exploration term into Eq. \ref{eq:performance_prob} based on bandit \cite{auer2002finite}. In machine learning, the multi-armed bandit problem is a classic reinforcement learning problem that exemplifies the exploration-exploitation trade-off dilemma: shall we stick to an arm that gave high reward so far (exploitation) or rather probe other arms further (exploration)? The Upper Confidence Bound (UCB) is widely used for dealing with the exploration-exploitation dilemma in the multi-armed bandit problem. Then, with the above analysis, Eq. \ref{eq:performance_prob} becomes
	\begin{equation}\label{eq:1-bit}
	    s(o^{(i,j)}_k) \leftarrow s(o^{(i,j)}_k) +  \delta *  \sqrt{\frac{2\log N}{n_{k,t}^{(i,j)}}} 
	\end{equation}
	where $N$ is the total number of samples, $n_{k,t}^{(i,j)}$ refers to the number of times the $k$th operation of edge $(i,j)$ has been selected, and $t$ is the index of the epoch. The first item in Eq.~\ref{eq:1-bit} is the value term which favors the operations that look good historically and the second is the exploration term which allows operations to get an exploration bonus that grows with $\log N$. And in this work $\delta = 2$ is used to balance value term and exploration term. We also test other values, which achieve a littler worse results. In that, 1-bit convolutions which behave badly in sufficient trials are prone to be abandoned. 
	
	Finally, we abandon the operation with the minimal selection likelihood for each edge. Such that the search space size is significantly reduced from $2 \times |\mathcal{O}^{(i,j)}|^{14}$ to $2 \times (|\mathcal{O}^{(i,j)}|-1)^{14}$.  We have
	
	\begin{equation}\label{eq:min}
		\mathcal{O}^{(i,j)} \leftarrow \mathcal{O}^{(i,j)} - \{ \mathop{\arg\min}\limits_{o^{(i,j)}_k}{s(o^{(i,j)}_k)} \}.
	\end{equation}
	
	The optimal structure is obtained when there is only one operation left in each edge. Our performance-based search algorithm is presented in Algorithm \ref{alg:MDL}. Note that in line 1, PC-DARTS is performed for $L$ epochs as the warm-up to find an initial architecture, and line 14 is used to update the architecture parameters $\alpha_{o^{(i,j)}_k}$ for all the edges due to the reduction of the search space $\{\mathcal{O}^{(i,j)}\}$.
	
	\begin{algorithm}[h]
		\caption{Performance-Based Search \label{alg:MDL}}
		\LinesNumbered
		\KwIn{Training data, Validation data, Searching hyper-graph: $\mathcal{G}$, $K=8$, {$T=3$, $V=1$, $L=5$}, $s(o^{(i,j)}_k)=0$ for all edges;}
		\KwOut{Optimal structure $\alpha$;}
		Search an architecture for $L$ epochs based on $\mathcal{O}^{(i,j)}$ using PC-DARTS; \\
		\While{$(K>1)$}{
			Select $\mathcal{O}^{(i,j)}_{smaller}$ consisting of $\lceil K/2 \rceil$ operations with smallest $\alpha_{o^{(i,j)}_k}$ from $\mathcal{O}^{(i,j)}$ for every edge; \\
			\For{$t= 1,...,T$ \rm epoch}{
				$\mathcal{O}^{'(i,j)}_{smaller} \leftarrow \mathcal{O}^{(i,j)}_{smaller}$; \\
				\For{ $ e=1,..., \lceil K/2 \rceil $ \rm epoch}{
					Select an architecture by sampling (without replacement) one operation from $\mathcal{O}^{'(i,j)}_{smaller}$ for every edge; \\
					Train the selected architecture and get the accuracy on the validation data; \\
					Assign this accuracy to all the sampled operations; \\
				}
			}
			Update $s(o^{(i,j)}_k)$ using Eq. \ref{eq:performance_prob}; \\
			\If{1 bit}{
				Update $s(o^{(i,j)}_k)$ using Eq. \ref{eq:1-bit}; \\
			}
			Update the search space \{$\mathcal{O}^{(i,j)}$\} using Eq. \ref{eq:min};\\
			Search the architecture for $V$ epochs based on $\mathcal{O}^{(i,j)}$ using PC-DARTS; \\
			$K = K -1$; \\
		}
	\end{algorithm}
	

	\subsection{{Gradient Update for BNAS}}
	
	%
		In BNAS, ${\hat X}^l$ in the $l^{th}$ layer are used to calculated the output feature maps $F^{l+1}$ as
		\begin{equation}
		  F^{l+1}=ACconv(F^l , \hat{X}^l),
		\end{equation}
	where $ACconv$ denotes the designed amplitude convolution operation in Eq.\ref{F}. 
	In ACconv, the channels of the output feature maps are generated as follows

		\begin{equation}
		  F^{l+1}_{h}=\sum_{i,g}F^l_g \otimes \hat{X}_{i}^l,
		  \label{F}
		\end{equation}
	where $\otimes$ denotes the convolution operation; $F^{l+1}_{h}$ is the $h^{th}$ feature map in the $({l+1})^{th}$ convolutional layer; $F^l_g$ denotes the $g^{th}$ feature map in the $l^{th}$ convolutional layer.
	{Note that the kernels of BNAS are binarized, while for 1-bit BNAS, both the kernels and the activations are binarized. Similar to the previous work \cite{rastegari2016xnor,liu2018bi,gu2018projection}, 
	the 1-bit BNAS is obtained via binarizing the kernels and activations simultaneously. 
	In addition, we replace ReLU with PReLU to reserve negative elements  generated by 1-bit convolution. }

	
		In BNAS, what need to be learned and updated are the full-precision kernels $X_i$ and amplitude matrices $A$. The kernels and the matrices are jointly learned. In each convolutional layer, BNAS update the full-precision kernels and then the amplitude matrices. In what follows, the layer index $l$ is omitted for simplicity.
	

			We denote $\delta_{X_i}$ as the gradient of the full-precision kernel $X_i$, and have
			
			\begin{equation}
				\delta_{X_i} = \frac{\partial L}{\partial X_i} = \frac{\partial L_S}{\partial X_i} + \frac{\partial L_{\hat A}}{\partial X_i},
				\label{delta_hatX}
			\end{equation}
			
			\begin{equation}
				X_i\leftarrow X_i - \eta_1 \delta _{X_i},
				\label{X_updating}
			\end{equation}
			where $\eta_1$ is a learning rate. We then have
			
			\begin{equation}
				\frac{\partial L_S}{\partial X_i} = \frac{\partial L_S}{\partial \hat{X}_i} \cdot \frac{\partial {\hat X}_i}{\partial X_i}= \frac{\partial L_S}{\partial \hat{X}_{i}} \cdot \hat A \cdot \mathds 1,
			\end{equation}
			
			\begin{equation}
				\frac{\partial L_{\hat A}}{\partial {X}_i}=\theta \cdot (X_i - \hat A \odot D_i),
			\end{equation}
			where $X_i$ is the full-precision convolutional kernel corresponding to $D_i$, and $\mathds{1}$ is the indicator function \cite{rastegari2016xnor} widely used to estimate the gradient of non-differentiable function.

			After updating $X$, we update the amplitude matrix $A$. Let $\delta_{A}$ be the gradient of ${A}$. According to Eq.\ref{loss_L}, we have
			
			\begin{equation}
			  \delta_{ A} = \frac{\partial L}{\partial {A}} = \frac{\partial L_S}{\partial {A}} + \frac{\partial L_{\hat A}}{\partial {A}},
			  \label{delta_A}
			\end{equation}
			
			\begin{equation}
			  A\leftarrow |A-\eta_2 \delta _{A}|,
			  \label{A_updating}
			\end{equation}
			where $\eta_2$ is another learning rate. Note that the amplitudes are always set to be non-negative. We then have
			\begin{equation}
				\frac{\partial L_S}{\partial {A}} = \sum_i\frac{\partial L_S}{\partial {\hat X}_i} \cdot \frac{\partial {\hat X}_i}{\partial \hat A} \cdot \frac{\partial \hat A}{\partial {A}}= \sum_i\frac{\partial L_S}{\partial \hat X_{i}} \cdot D_i,
			\end{equation}
			
			\begin{equation}
				\frac{\partial L_{\hat A}}{\partial {A}}=\frac{\partial L_{\hat A}}{\partial \hat A} \cdot \frac{\partial \hat A}{\partial {A}}=-\theta \cdot (X_i - \hat A \odot D_i) \cdot D_i,
				\label{delta_LA_A}
			\end{equation}
			where $\frac{\partial \hat A}{\partial {A}}$ is set to $1$ for easy implementation of the algorithm. Note that $\hat A$ and $A$ are respectively used in the forward pass and the back propagation in an asynchronous manner. The above derivations show that BNAS is learnable with the new BP algorithm. 

	\section{Experiments}\label{sec:experiment}
	
	In this section, we compare our BNAS with state-of-the-art NAS methods, and also validate two BNAS models based on XNOR \cite{rastegari2016xnor} and PCNN \cite{gu2018projection}. The 1-bit BNAS models are also included in our experiments to further validate our methods.

	\subsection{Experiment Protocol}
	
	\subsubsection{Datasets}
	
	\textbf{CIFAR-10:} CIFAR-10 \cite{krizhevsky2014cifar} is a natural image classification dataset, which is composed of a training set and a test set, with 50,000 and 10,000 32$\times$32 color images, respectively. These images span 10 different classes, including airplanes, automobiles, birds, cats, deer, dogs, frogs, horses, ships and trucks.

	\textbf{ILSVRC12 ImageNet:} ILSVRC12 ImageNet object classification dataset \cite{russakovsky2015imagenet} is more diverse and challenging. It contains 1.2 million training images, and 50,000 validation images, across 1000 classes.
	
	
	\textbf{CASIA-WebFace:} CASIA-WebFace \cite{dongyi2014learning} is a face image dataset collected from over ten thousand different individuals, containing nearly half a million facial images. Note that compared to other private datasets used in DeepFace \cite{Taigman2014DeepFace} (4M), VGGFace \cite{parkhi2015deep} (2M) and FaceNet \cite{schroff2015facenet} (200M), our training data contains just 490K images and is more challenging. 
	
	\textbf{LFW:} The Labeled Faces in the Wild (LFW) dataset \cite{huang2008labeled} has 5,749 celebrities and collected 13,323 photos of them from web. The photos are organized into 10 splits, each of which contain 6000 images. Celebrities in Frontal-Profile (CFP) \cite{sengupta2016frontal} consists of 7000 images of 500 subjects. The dataset contains 5000 images in frontal view and 2000 images in extreme profile to evaluate the performance on coping with the pose variation. The data is divided into 10 splits, each containing an equal number of frontal-frontal and frontal-profile comparisons. 
	
	\textbf{AgeDB:} AgeDB \cite{moschoglou2017agedb} includes 16,488 images of various famous people. The images are categorized to 568 distinct subjects according to their identity, age and gender attributes.

	\subsubsection{Train and Search Detials}
	In these experiments, we first search neural architectures on an over-parameterized network on CIFAR-10, and then evaluate the best architecture with a stacked deeper network on the same dataset. Then we further perform experiments to search architectures directly on ImageNet. We run the experiment multiple times and find that the resulting architectures only show slight variation in performance, which demonstrates the stability of the proposed method.
	
	\begin{table}[htbp]
        \caption{Test error rates for human-designed full-precision networks, human-designed binarized networks, full-precision networks obtained by NAS, and networks obtained by our BNAS on CIFAR-10. 'W' and 'A' refer to the weight and activation bitwidth respectively. For fair comparison, we select the architectures by NAS with similar parameters ($<$ $5$M). In addition, we also train an optimal architecture in a larger setting, \emph{i.e.,} with more initial channels ($44$ in XNOR or $48$ in PCNN). {$\dag$ Indicate that BNAS is performed based on Eq. \ref{eq:1-bit}, which is also the same case in the following experiments.} {$*$ Indicate that the result is tested by the quantized NAS architecture obtained by PC-DARTS.}}
		\begin{center}
			\resizebox{\textwidth}{43mm}{
				\begin{tabular}{lcccccc}
					\toprule[1pt]
					\multirow{2}{*}{\textbf{Architecture}} & \textbf{Test Error} & \textbf{\# Params}  & \multirow{2}{*}{\textbf{W}} & \multirow{2}{*}{\textbf{A}} & \textbf{Search Cost} & \textbf{Search} \\
					& \textbf{(\%)} & \textbf{(M)} && & \textbf{(GPU days)} & \textbf{Method} \\
					\hline
					ResNet-18 \cite{he2016deep}  & 3.53  & 11.1 & 32 & 32 & - & Manual \\
					WRN-22 \cite{zagoruyko2016wide}  & 4.25 & 4.33 & 32 & 32 & - & Manual  \\
					DenseNet \cite{huang2017densely} & 4.77 & 1.0 & 32 & 32 & - & Manual \\
					SENet \cite{hu2018squeeze} & 4.05 & 11.2 & 32 & 32 & - & Manual \\
					NASNet-A \cite{Zoph2018CVPR} & 2.65 & 3.3 & 32 & 32 & 1800 & RL \\
					AmoebaNet-A \cite{real2019regularized} & 3.34 & 3.2 & 32 & 32 & 3150 & Evolution \\
					PNAS \cite{liu2018progressive} & 3.41 & 3.2 & 32 & 32 & 225 & SMBO \\
					ENAS \cite{pham2018efficient}  & 2.89 & 4.6 & 32 & 32 & 0.5 & RL \\
					Path-level NAS \cite{cai2018path} & 3.64 & 3.2 & 32 & 32 & 8.3 & RL \\
					DARTS(first order) \cite{liu2018darts}  & 2.94 & 3.1 & 32 & 32 & 1.5 & Gradient-based \\
					DARTS(second order) \cite{liu2018darts} & 2.83 & 3.4 & 32 & 32 & 4 & Gradient-based \\
					PC-DARTS & 2.78 & 3.5 & 32 & 32 & 0.15 & Gradient-based \\ 
					\textbf{BNAS} (full-precision) & 2.84  & 3.3 & 32 & 32 & 0.08 & Performance-based \\					
					\hline
					Network in {\cite{mcdonnell2018training}} & 6.13 & 4.30 & 1 & 32 & - & Manual\\
					ResNet-18 (XNOR) & 6.69 & 11.17 & 1 & 32  & - & Manual \\
					ResNet-18 (PCNN) & 5.63 & 11.17 & 1 & 32  & - & Manual \\
					WRN22 (PCNN) \cite{gu2018projection} & 5.69 & 4.29 & 1 & 32  & - & Manual \\
					{PC-DARTS$^*$} & 4.86 & 3.638 & 1 & 32 & 0.15 & Gradient-based \\
					{PC-DARTS} & 4.88 & 3.1 & 1 & 32 & 0.18 & Gradient-based \\ %
					\textbf{BNAS} (XNOR) & 5.71 & 2.3 & 1 & 32 & 0.104 & Performance-based \\
					\textbf{BNAS} (XNOR, larger) & 4.88 & 3.5 & 1 & 32 & 0.104 & Performance-based \\
					\textbf{BNAS}  & 3.94 & 2.6 & 1 & 32 & 0.09375 & Performance-based \\
					{\textbf{BNAS}$^{\dag}$} & 4.01 & 2.7 & 1 & 32 & 0.094 & Performance-based \\
					\textbf{BNAS} (larger) & 3.47 & 4.6 & 1 & 32 & 0.09375 & Performance-based \\
					\hline
					ResNet-18 (PCNN) \cite{liu2019rbcn} & 14.5 & 0.59 & 1 & 1  & - & Manual \\
					WRN22 (XNOR) \cite{zhao2019bonn} & 11.48 & 4.33 & 1 & 1  & - & Manual \\
					WRN22 (PCNN) \cite{liu2019rbcn} & 8.38 & 2.4 & 1 & 1 & - & Manual \\
					{PC-DARTS} & 8.94 & 4.2 & 1 & 1 & 0.21 & Gradient-based \\ %
					\textbf{BNAS} & 8.3 & 4.6 & 1 & 1 & 0.112 & Performance-based \\
					\textbf{BNAS}$^{\dag}$ & 6.72 & 4.7 & 1 & 1 & 0.113 & Performance-based \\
					\bottomrule[1pt]
				\end{tabular}}
			\end{center}
			
			\label{tab:cifar_results}
		\end{table}

		We use the same datasets and evaluation metrics as existing NAS works \cite{liu2018darts,cai2018path,Zoph2018CVPR,liu2018progressive}. First, most experiments are conducted on CIFAR-10 \cite{krizhevsky2009learning}, 
		and the color intensities of all images are normalized to $[-1, +1]$. During architecture search, the $50$K training samples of CIFAR-10 is divided into two subsets of equal size, one for training the network weights and the other for searching the architecture hyper-parameters. When reducing the search space, we randomly select $5$K images from the training set as a validation set (used in line 8 of Algorithm \ref{alg:MDL}). 
		Specially for 1-bit BNAS, we replace ReLU with PReLU to avoid the disappearance of negative numbers generated by 1-bit convolution, and the bandit strategy is introduced to solve the insufficient training problem caused by the binarization of both kernels and activations.
		To further show the efficiency of our method, we also search architecture on ImageNet directly.
		
		In the search process, we consider a total of $6$ cells in the network, where the reduction cell is inserted in the second and the fourth layers, and the others are normal cells. There are $M=4$ intermediate nodes in each cell. Our experiments follow PC-DARTS. We set the hyper-parameter $C$ in PC-DARTS to $2$ for CIFAR-10 so only $1/2$ features are sampled for each edge. The batch size is set to $128$ during the search of an architecture for $L=5$ epochs based on $\mathcal{O}^{(i,j)}$ (line 1 in Algorithm \ref{alg:MDL}). {Note that for $5 \le L \le 10$, a larger $L$ has little effect on the final performance, but costs more search time as shown in Tab. \ref{table:L}.} We freeze the network hyper-parameters such as $\alpha$, and only allow the network parameters such as filter weights to be tuned in the first $3$ epochs.  Then in the next 2 epochs, we train both the network hyper-parameters and the network parameters. This is to provide an initialization for the network parameters and thus alleviates the drawback of parameterized operations compared with free parameter operations. We also set $T = 3$ (line 4 in Algorithm \ref{alg:MDL}) and $V = 1$ (line 14), so the network is trained less than $60$ epochs, with a larger batch size of $400$ (due to few operation samplings) during reducing the search space. The initial number of channels is $16$. We use SGD with momentum to optimize the network weights, with an initial learning rate of $0.025$ (annealed down to zero following a cosine schedule), a momentum of 0.9, and a weight decay of $5 \times 10^{-4}$. The learning rate for finding the hyper-parameters is set to $0.01$. When we search architecture directly on ImageNet, we use the same parameters with searching on CIFAR-10 except that initial learning rate is set to $0.05$

    	\begin{table}[t]
    		\centering
    		\caption{{With different $L$, the accuracy and search cost of BNAS based on PCNN on CIFAR10 dataset.}}
    		\begin{tabular}{c c c c c c}
    			\toprule
    			\multirow[c]{2}{*}{Model} & \multicolumn{5}{c}{$L$} \\
    			\cmidrule(lr){2-6}
    			& $3$ & $5$ & $7$ & $9$ & $11$ \\
    			\hline
    			\hline
    			Accuracy (\%) & 95.8 & 96.06 & 95.94 & 96.01 & 96.03 \\
    			Search cost & 0.0664 & 0.09375 & 0.1109 & 0.1321 & 0.1687 \\
    			\bottomrule
    		\end{tabular}
    		
    		\label{table:L}
    	\end{table}
	
		After search, in the architecture evaluation step, our experimental setting is similar to \cite{liu2018darts,Zoph2018CVPR,pham2018efficient}. A larger network of $20$ cells ($18$ normal cells and $2$ reduction cells) is trained on CIFAR-10 for $600$ epochs with a batch size of $96$ and an additional regularization cutout \cite{devries2017improved}. The initial number of channels is $36$. We use the SGD optimizer with an initial learning rate of $0.025$ (annealed down to zero following a cosine schedule without restart), a momentum of $0.9$, a weight decay of $3 \times 10^{-4}$ and a gradient clipping at $5$. When stacking the cells to evaluate on ImageNet, the evaluation stage follows that of DARTS, which starts with three convolution layers of stride $2$ to reduce the input image resolution from $224 \times 224$ to $28 \times 28$. $14$ cells ($12$ normal cells and $2$ reduction cells) are stacked after these three layers, with the initial channel number being $64$. The network is trained from scratch for $250$ epochs using a batch size of $512$. We use the SGD optimizer with a momentum of $0.9$, an initial learning rate of $0.05$ (decayed down to zero following a cosine schedule), and a weight decay of $3 \times 10^{-5}$. Additional enhancements are adopted including label smoothing and an auxiliary loss tower during training. All the experiments and models are implemented in PyTorch \cite{paszke2017automatic}.
		
				\begin{figure}[h]
					\centering
					\subfigure[Normal Cell]{
						\includegraphics[scale=.45]{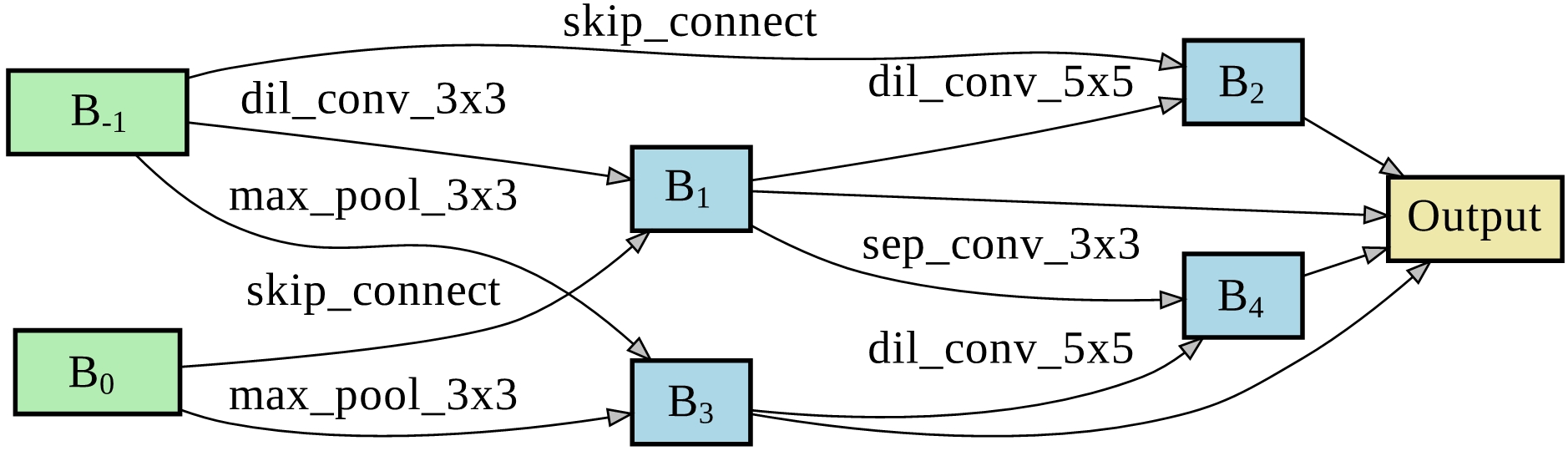}
					}
					\quad
					
					\subfigure[Reduction Cell]{
						\includegraphics[scale=.3]{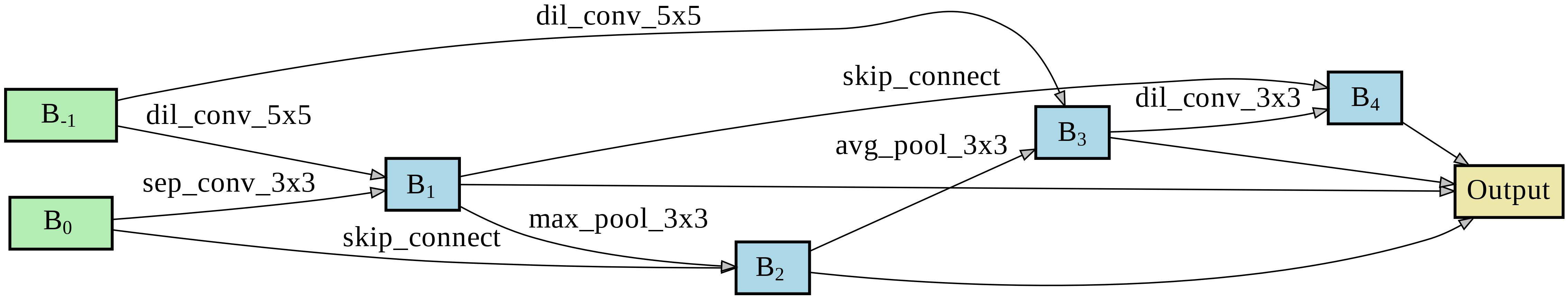}
					}
					\caption{Detailed structures of the best cells discovered on CIFAR-10 using BNAS based on XNOR. In the normal cell, the stride of the operations on $2$ input nodes is 1, and in the reduction cell, the stride is 2.}
					\label{fig:xnor}
				\end{figure}
			\noindent

		\subsection{Results on CIFAR-10}
		We compare our method with both manually designed networks and networks searched by NAS. The manually designed networks include ResNet \cite{he2016deep}, Wide ResNet (WRN) \cite{zagoruyko2016wide}, DenseNet \cite{huang2017densely} and SENet \cite{hu2018squeeze}. For the networks obtained by NAS, we classify them according to different search methods, such as RL (NASNet \cite{Zoph2018CVPR}, ENAS \cite{pham2018efficient}, and Path-level NAS \cite{cai2018path}), evolutional algorithms (AmoebaNet \cite{real2019regularized}), Sequential Model Based Optimization (SMBO) (PNAS \cite{liu2018progressive}), and gradient-based methods (DARTS \cite{liu2018darts} and PC-DARTS \cite{xu2019pcdarts}).
		
			\begin{figure}[h]
				\centering
				\subfigure[Normal Cell]{
					\includegraphics[scale=.45]{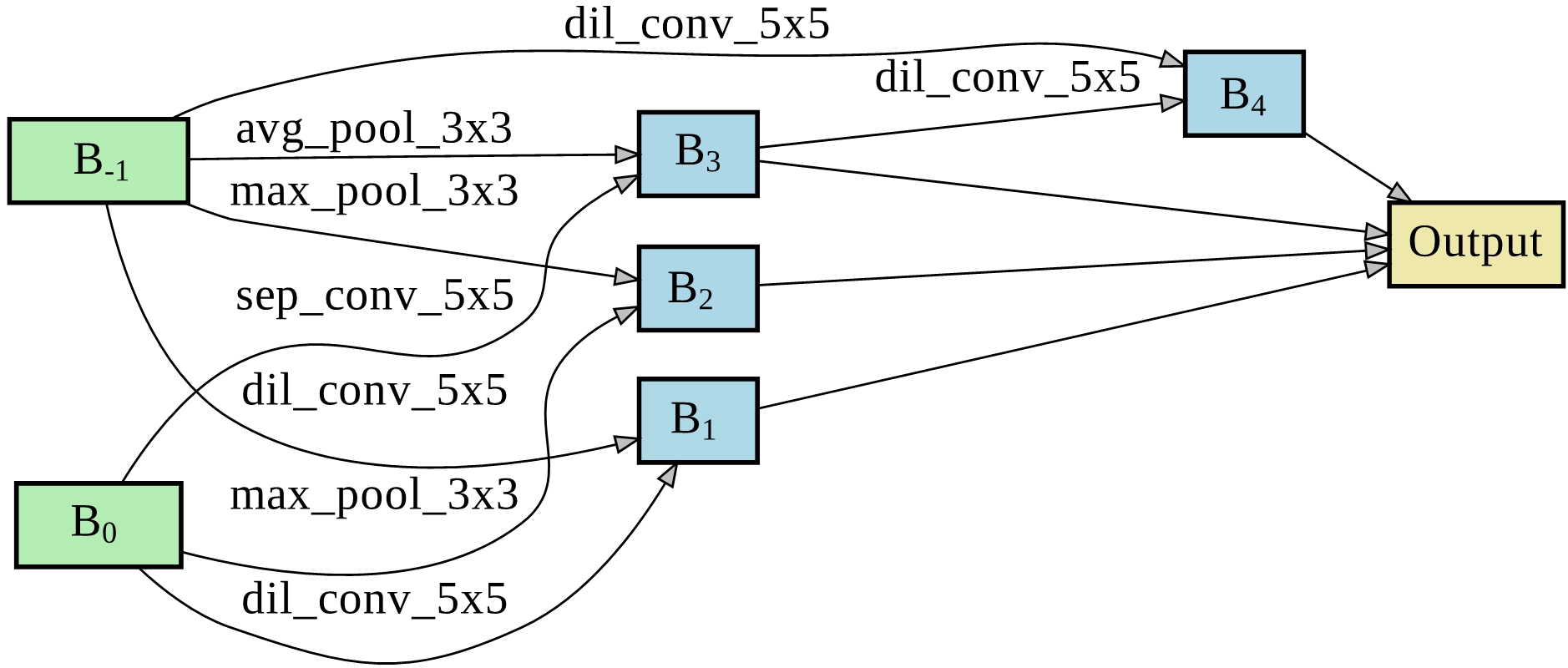}
				}
				\quad
				\subfigure[Reduction Cell]{
					\includegraphics[scale=.4]{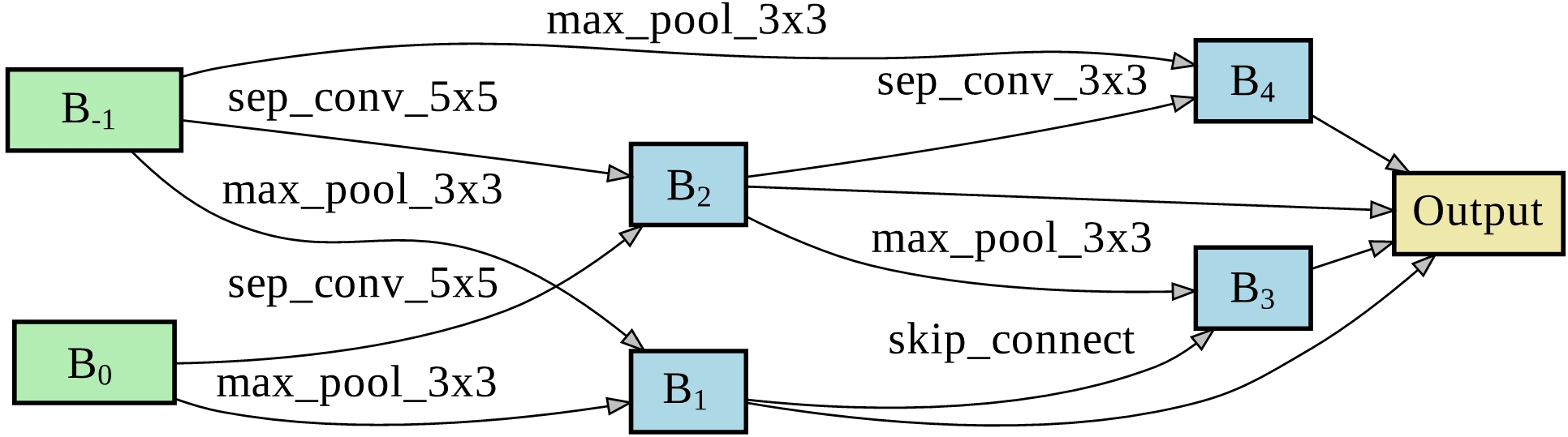}
				}
				\caption{Detailed structures of the best cells discovered on CIFAR-10 using BNAS based on PCNN. In the normal cell, the stride of the operations on $2$ input nodes is 1, and in the reduction cell, the stride is 2.}
				\label{fig:mcn}
			\end{figure}
			
		The results for different architectures on CIFAR-10 are summarized in Tab. \ref{tab:cifar_results}. Using BNAS, we search for two binarized networks based on XNOR \cite{rastegari2016xnor} and PCNN \cite{gu2018projection}. In addition, we also train a larger XNOR variant with $44$ initial channels and a larger PCNN variant with $48$ initial channels. We can see that the test errors of the binarized networks obtained by our BNAS are comparable to or smaller than those of the full-precision human designed networks, and are significantly smaller than those of the other binarized networks.

		Compared with the full-precision networks obtained by other NAS methods, the binarized networks by our BNAS have comparable test errors but with much more compressed models. Note that the numbers of parameters of all these searched networks are less than 5M, but the binarized networks only need $1$ bit to save one parameter, while the full-precision networks need $32$ bits. {For 1-bit BNAS, as shown in Tab. \ref{tab:cifar_results}, the UCB improves it by $1.58\%$, which validates the effectiveness of our method. Also, we observe that up to $1.66\%$ accuracy improvement is gained with 1-bit BNAS.} In terms of search efficiency, compared with the previous fastest PC-DARTS, our BNAS is $40\%$ faster (tested on our platform (NVIDIA GTX TITAN Xp). We attribute our superior results to the proposed way of solving the problem with the novel scheme of search space reduction. As illustrated in Figs. \ref{fig:xnor} and \ref{fig:mcn}, compared with NAS, the architectures of BNAS prefer larger receptive fields. It also results in more pooling operations, most of which can increase the nonlinear representation ability of BNNs.
		
		Our BNAS method can also be used to search full-precision networks. In Tab. \ref{tab:cifar_results}, BNAS (full-precision) and PC-DARTS perform equally well, but BNAS is $47\%$ faster. Both the binarized methods XNOR and PCNN in our BNAS perform well, which shows the generalization of BNAS. Fig. \ref{fig:xnor} and Fig. \ref{fig:mcn} show the best cells searched by BNAS based on XNOR and PCNN, respectively.
		
		We also use PC-DARTS to perform a binarized architecture search based on PCNN on CIFAR10, resulting in a network denoted as PC-DARTS (PCNN). Compared with PC-DARTS (PCNN), BNAS  achieves a better performance ($95.12$\% vs. $96.06$\% in test accuracy) with less search time ($0.18$ vs. $0.09375$ GPU days). {We also compare our 1-bit BNAS with PC-DARTS, and find that our method is better than PC-DARTS ($93.28\%$ vs. $90.06\%$) on CIFAR-10 and about twice as fast as PC-DARTS ($0.113$ vs. $0.21$ GPU days).} The reason for this may be because the performance based strategy can help find better operations for recognition.

		\begin{table}[htbp]
            \caption{Comparison with the state-of-the-art image classification methods on ImageNet. 'W' and 'A' refer to the weight and activation bitwidth respectively. BNAS and PC-DARTS are obtained directly by NAS and BNAS on ImageNet, others are searched on CIFAR-10 and then directly transferred to ImageNet.}
			\begin{center}
				\resizebox{\textwidth}{25mm}{
					\begin{tabular}{lccccccc}
						\toprule
						\multirow{2}{*}{\textbf{Architecture}} & \multicolumn{2}{c}{\textbf{Accuracy (\%)}} & \textbf{Params} & \multirow{2}{*}{\textbf{W}} & \multirow{2}{*}{\textbf{A}} & \textbf{Search Cost} & \textbf{Search} \\ \cline{2-3}
						& \textbf{Top1} & \textbf{Top5} & \textbf{(M)} && & \textbf{(GPU days)} & \textbf{Method} \\ 
						\hline
						ResNet-18 \cite{gu2018projection} & 69.3 & 89.2 & 11.17 & 32 & 32 & - & Manual \\
						MobileNetV1 \cite{howard2017mobilenets} & 70.6 & 89.5 & 4.2 & 32 & 32 & - & Manual \\
						\hline
						NASNet-A \cite{Zoph2018CVPR} & 74.0  & 91.6 & 5.3 & 32 & 32 & 1800 & RL \\
						AmoebaNet-A \cite{real2019regularized} & 74.5 & 92.0 & 5.1 & 32 & 32 & 3150 & Evolution \\
						AmoebaNet-C \cite{real2019regularized} & 75.7 & 92.4 & 6.4 & 32 & 32 & 3150 & Evolution \\
						PNAS \cite{liu2018progressive} & 74.2 & 91.9 & 5.1 & 32 & 32 & 225 & SMBO \\
						DARTS \cite{liu2018darts} & 73.1 & 91.0 & 4.9 & 32 & 32 & 4 & Gradient-based \\
						PC-DARTS \cite{xu2019pcdarts} & 75.8 & 92.7 & 5.3 & 32 & 32 & 3.8 & Gradient-based \\
						\hline
						ResNet-18 (PCNN) \cite{gu2018projection} & 63.5 & 85.1 & 11.17 & 1 & 32 & - & Manual \\
						\textbf{BNAS} & 71.3 & 90.3 & 6.2 & 1 & 32 & 2.6 & Performance-based \\

						\hline
						
						ResNet-18 (Bi-Real) \cite{liu2018bi} & 56.4 & 79.5 & 11.17 & 1 & 1 & - & Manual \\
						ResNet-18 (BONN) \cite{zhao2019bonn} & 59.3 & 81.6 & 11.17 & 1 & 1 & - & Manual \\
						ResNet-18 (PCNN) \cite{gu2018projection} & 57.3 & 80.0 & 11.17 & 1 & 1 & - & Manual \\
                        \textbf{BNAS}  & 64.3 & 86.1 & 6.4 & 1 & 1 & 3.2 & Performance-based \\
						
						\bottomrule
					\end{tabular}}
				\end{center}
				
				\label{tab:imagenet_results}
			\end{table}

			\subsection{Results on ImageNet}
\begin{figure}[h]
				\centering
				\subfigure[Normal Cell]{
					\includegraphics[scale=.45]{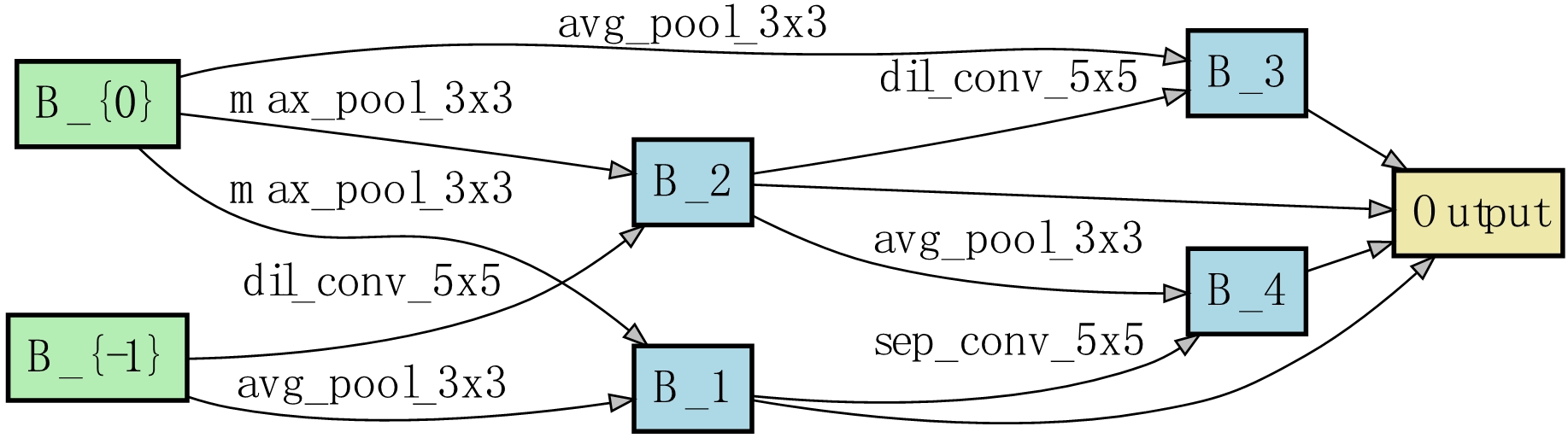}
				}
				\quad
				\subfigure[Reduction Cell]{
					\includegraphics[scale=.4]{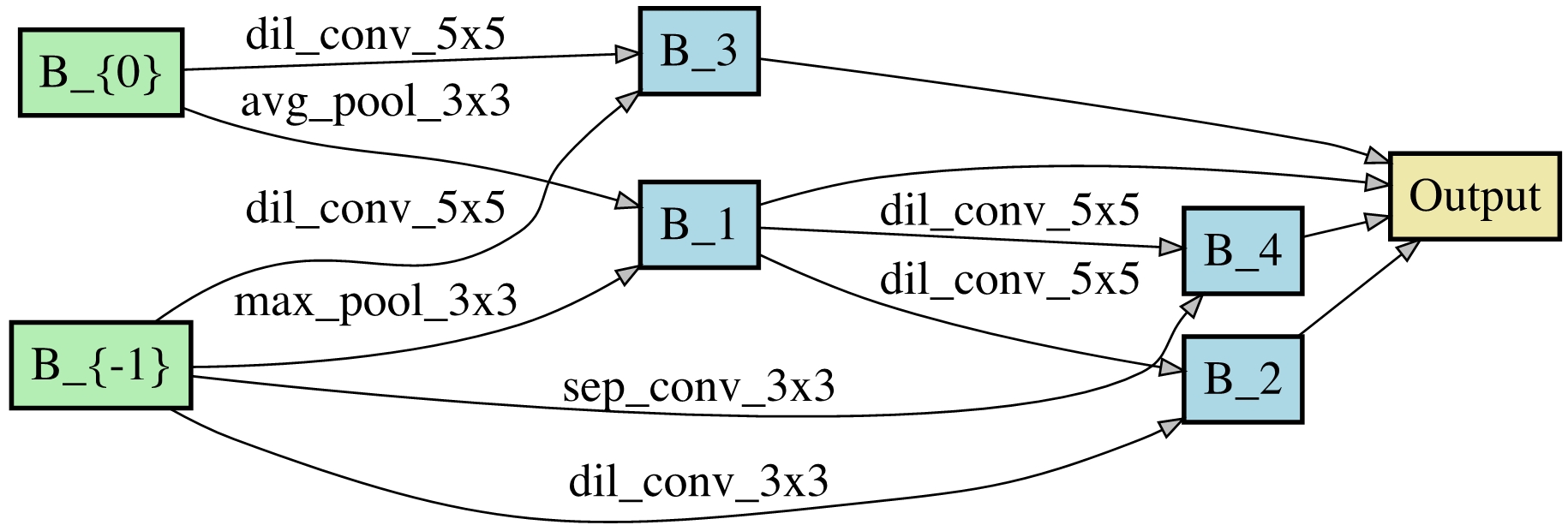}
				}
				\caption{Detailed structures of the best cells discovered on ImageNet using BNAS based on PCNN. In the normal cell, the stride of the operations on $2$ input nodes is 1, and in the reduction cell, the stride is 2.}
				\label{fig:imagenet}
			\end{figure}

			We further compare the state-of-the-art image classification methods on ImageNet. All the searched networks are obtained directly by NAS and BNAS on ImageNet by stacking the cells. Due to the large number of categories and data, ImageNet is more challenging than CIFAR-10 for binarized network. Different from the architecture settings for CIFAR-10, we do not binarize  the first convolutional layer in depth-wise separable convolution and the preprocessing operations for 2 input nodes. Instead, we replace the concatenation with summation for the preprocessing operations and increase the number of channels for each cell. The benefits are more focusing on model compression with the state-of-the-art performance. 
            From the results in Tab. \ref{tab:imagenet_results}, we have the following observations: (1) BNAS  performs better than human-designed binarized networks { (71.3\% vs. 63.5\%)} and has far fewer parameters { (6.1M vs. 11.17M)}. 
			(2) BNAS has a performance similar to the human-designed full-precision light networks { (71.3\% vs. 70.6\%)},  with a much more highly compressed model. 
			(3) 1-bit BNAS achieves $5.0\%$ accuracy improvement than the state-of-the-art human-designed 1-bit network, with fewer parameters.
			(4) Compared with the full-precision networks obtained by other NAS methods, BNAS has little performance drop, but is fastest in terms of search efficiency { (0.09375 vs. 0.15 GPU days)} and is a much more highly compressed model due to the binarization of the network. The above results show the excellent transferability of our BNAS method. \ref{fig:imagenet} shows the best cells searched by BNAS based on PCNN. They perform comparably to the full-precision networks obtained by NAS methods, but with highly compressed models.
			
	\subsection{Results on Face Recognition}
	
	\begin{table}[tbp]
		\caption{Test accuracies based on ResNet-18, ResNet-34, ResNet-50, ResNet-100 and BNAS on face recognition datasets. 'W' and 'A' refer to the weight and activation bitwidth respectively. We train these models on the CASIA-WebFace dataset, but the test process are performed on the following datasets: LFW, CFP, AgeDB. On all the three test datasets, the results of BNAS consistently outperform the other methods. }
		\centering
			\resizebox{\textwidth}{23mm}{
			\begin{tabular}{lcccccccc}
				\toprule[1pt]
				\multirow{2}{*}{\textbf{Architecture}} & \multicolumn{3}{c}{\textbf{Accuracy(\%)}} &  \textbf{\# Params} & \multirow{2}{*}{\textbf{W}} & \multirow{2}{*}{\textbf{A}} & \textbf{Search Cost} & \textbf{Search} \\
				& \textbf{LFW} & \textbf{CFP} & \textbf{AgeDB} & \textbf{(M)} & & & \textbf{(GPU days)} & \textbf{Method} \\
				\hline
				ResNet-18 \cite{he2016deep} & 98.68 & 92.33 & 90.23 & 24.02 & 32 & 32 & - & Manual \\
				ResNet-34 \cite{he2016deep} & 99.03 & 92.98 & 91.15 & 36.56 & 32 & 32 & - & Manual \\
				ResNet-50 \cite{he2016deep} & 99.07 & 93.73 & 91.58 & 45.46 & 32 & 32 & - & Manual \\
				ResNet-100 \cite{he2016deep} & 99.20 & 92.22 & 93.99 & 75.58 & 32 & 32 & - & Manual \\
				\hline
				ResNet-18 (XNOR) & 92.03 & 75.04 & 72.13 & 24.02 & 1 & 1 & - & Manual \\
				ResNet-18 (PCNN) & 94.32 & 80.01 & 77.55 & 24.02 & 1 & 1 & - & Manual \\
				ResNet-34 (XNOR) & 91.65 & 73.94 & 71.98 & 36.56 & 1 & 1 & - & Manual \\
				ResNet-34 (PCNN) & 94.58 & 80.59 & 77.50 & 36.56 & 1 & 1 & - & Manual \\
				ResNet-50 (XNOR) & 92.03 & 75.50 & 72.12 & 45.46 & 1 & 1 & - & Manual \\
				ResNet-100 (XNOR) & 92.34 & 75.01 & 72.80 & 75.58 & 1 & 1 & - & Manual \\
				\hline
				\textbf{BNAS} & 98.57 & 92.46 & 89.03 & 10.224 & 1 & 32 & 0.717 & Performance-based \\
				\textbf{BNAS} & 97.62 & 89.89 & 83.6 & 10.768 & 1 & 1 & 0.856 & Performance-based \\
				\bottomrule[1pt]
			\end{tabular}}
		\label{table:face}
	\end{table}
	
	In this section, we compare different kinds of ResNets with BNAS on face recognition task. Different kinds of ResNets are ResNet-18, ResNet34, ResNet-50 and ResNet-100 with kernel stage, $64$-$128$-$256$-$512$ and each model has two FC layers. We directly search on CASIA-Webface for $17.2$ hours using one TITAN V GPU with $400$ batch size, learning rate of $0.05$. We use CASIA-Webface dataset for training and LFW, CFP, AgeDB datasets for testing. The setting of hyper-parameters is similar to the strategy of CIFAR experiments, despite the difference that the learning rate is $0.05$ and the maximum epochs is set to $100$. Note that the amount of parameters of ResNet is huge because we remove the pooling operation before FC layer following the face recognition code$\footnote{\url{https://github.com/wujiyang/Face_Pytorch}}$. It makes the fully connected layer parameters large.
	
	As demonstrated in Tab.\;\ref{table:face}, BNAS has a performance similar to the human-designed full-precision networks ResNet-18, with a much more highly compressed model. Also, 1-bit BNAS not only achieves the best test result among 1-bit CNNs but also has fewest parameters. On LFW, 1-bit BNAS has only $1.06\%$ accuracy degradation compared to the results of the full-precision models ResNet-18, which verify the potential of 1-bit networks in practice.

	\section{Conclusion}

	In this paper, we introduce BNAS (1-bit BNAS) for efficient object recognition, which is the first binarized neural architecture search algorithm. Our BNAS can effectively reduce the search time by pruning the search space in early training stages, which is faster than the previous most efficient search method PC-DARTS. We also introduce the bandit strategy into 1-bit BNAS, which can significantly improve the performance. The binarized networks searched by BNAS can achieve excellent accuracies on CIFAR-10, ImageNet, and wild face recognition. They perform comparably to the full-precision networks obtained by other NAS methods, but with much compressed models.

	
    \begin{acknowledgements}
    The work was supported in part by National Natural Science Foundation of China under Grants 61672079. This work is supported by Shenzhen Science and Technology Program KQTD2016112515134654. Baochang Zhang is also with Shenzhen Academy of Aerospace Technology, Shenzhen 100083, China. Hanlin Chen and Li'an Zhuo have the same contributions to the paper.
    \end{acknowledgements}

%
%

\bibliographystyle{spmpsci}      
\bibliography{bibliography}   

%
%

\end{document}